\documentclass[10pt,twocolumn,letterpaper]{article}

\pdfoutput=1
\usepackage{cvpr}
\usepackage{times}
\usepackage{graphicx}
\usepackage{amsmath}
\usepackage{amssymb}

\usepackage{booktabs}
\usepackage{multirow}
\usepackage[breaklinks=true,letterpaper=true,bookmarks=false]{hyperref}

\cvprfinalcopy 

\ifcvprfinal\fi
\begin{document}

\title{Unbiased Scene Graph Generation via Rich and Fair Semantic Extraction}

\author{Bin Wen, Jie Luo, Xianglong Liu\\
State Key Laboratory of Software Development Environment,\\
School of Computer Science and Engineering, Beihang University\\
No. 37, Xueyuan Road, Beijing 100191, P.R. China\\
{\tt\small \{wenbin,luojie,xlliu\}@nlsde.buaa.edu.cn}
\and
Lei Huang\\
Inception Institute of Artificial Intelligence (IIAI)\\
Abu Dhabi, UAE\\
{\tt\small lei.huang@inceptioniai.org}
}

\maketitle

\begin{abstract}
Extracting graph representation of visual scenes in image is a challenging task in computer vision. Although there has been encouraging progress of scene graph generation in the past decade, we surprisingly find that the performance of existing approaches is largely limited by the strong biases, which mainly stem from (1) unconsciously assuming relations with certain semantic properties such as symmetric and (2) imbalanced annotations over different relations. To alleviate the negative effects of these biases, we proposed a new and simple architecture named Rich and Fair semantic extraction network (RiFa for short), to not only capture rich semantic properties of the relations, but also fairly predict relations with different scale of annotations. Using pseudo-siamese networks, RiFa embeds the subject and object respectively to distinguish their semantic differences and meanwhile preserve their underlying semantic properties. Then, it further predicts subject-object relations based on both the visual and semantic features of entities under certain contextual area, and fairly ranks the relation predictions for those with a few annotations. Experiments on the popular Visual Genome dataset show that RiFa achieves state-of-the-art performance under several challenging settings of scene graph task. Especially, it performs significantly better on capturing different semantic properties of relations, and obtains the best overall per relation performance.
\end{abstract}


\section{Introduction}
Extracting explicit semantic representation from images is one of the main challenges in computer vision. Recently, there has been increasing interest on using scene graph \cite{Johnson2015} as representation, in which objects in the image are represented as vertices and relations between objects are represented as edges. From the perspective of knowledge, a scene graph can be viewed as a kind of knowledge graph, where objects are corresponding to entities, classes of objects are corresponding to classes of entities, and relations are corresponding to object properties between entities.

In the past decade, there has attracted a number of studies that attempted to generate the scene graph using various techniques. There are mainly two types of methods for scene graph generation. The first type concentrates on utilizing internal information such as the visual features of images. Typical techniques include iterative message passing \cite{xu2017scenegraph}, associative embeddings \cite{Newell2017Pixels}, graph neural network \cite{Yang2018,Wang2019}. The second type tries to combine internal information and external knowledge such as language features and knowledge graphs. Typical techniques include motifs \cite{Zellers2017Neural}, knowledge distillation \cite{Plesse2018Visual}, knowledge embedding \cite{Zhang2019}. Unfortunately, in practice these methods often learn the biased models for scene graph generation, mainly due to the incorrect assumption of relations properties and the imbalanced annotations in the training set.

\begin{figure}[t]
  \centering
  \includegraphics[width=\columnwidth]{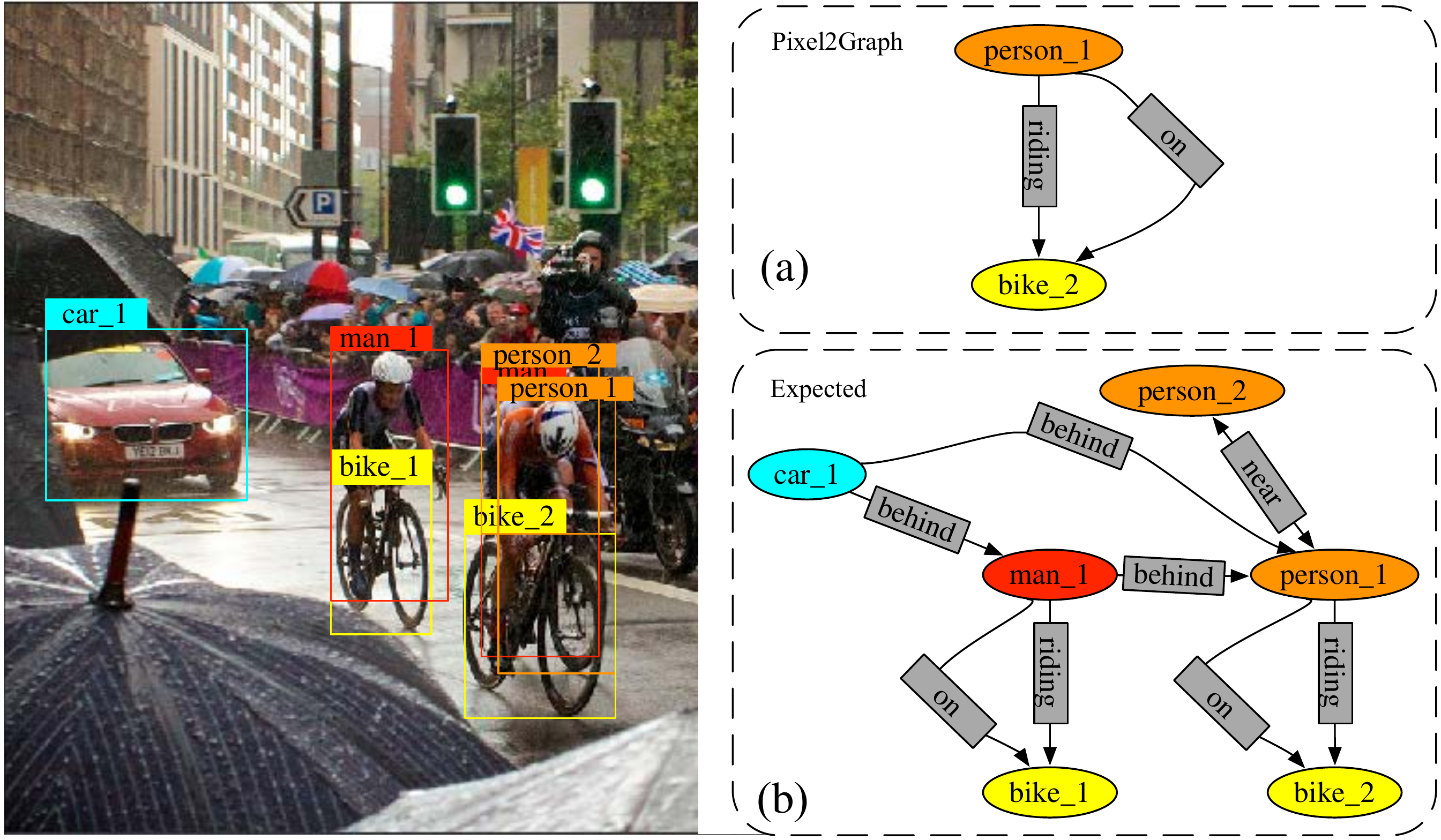}
  \caption{An example of relations with different properties. (a) is the scene graph generated by the Pixel2Graph model \cite{Newell2017Pixels}; although its prediction matches the ground truth, there are still many relations and their properties not being recognized as shown in (b). For example, relation ``riding'' is asymmetric, i.e. if (man\_1, riding, bike\_1) is a relation instance, then its symmetric form (bike\_1, riding, man\_1) is not a relation instance; relation ``near'' is symmetric, i.e. if (person\_1, near, person\_2) is a relation instance, then (person\_2, near, person\_1) is also a relation instance; relation ``behind'' is transitive, i.e. if (car\_1, behind, man\_1) and (man\_1, behind, person\_1) are both relation instance, then (car\_1, behind, person\_1) is also a relation instance. These semantic properties of relations are reflected by the triples that constitute the scene graph.}\label{fig:rel-ex}
\end{figure}

First, most of these methods treat the relations as the labels, usually based on the biased assumption that the relations among entities follow certain semantic properties like symmetry between the subject and object. In fact, the scene graph is a formal representation of knowledge, where relations are not just labels for classifying pairs of entities, but own specific semantic properties that serves as a meta constraint defining the exact meanings of the relations (see examples in Figure~\ref{fig:rel-ex}). Second, the imbalanced annotations for different relations lead the trained models to mainly focus on the frequent relations, and inevitably generates the biased prediction ignoring those uncommonly occurred ones, which may be very valuable in applications such as medical diagnosis. The imbalanced relations can be often observed in the widely used datasets for scene graph task \cite{Chen2019,Zhang2019}, such as Visual Genome.

To avoid the two kinds of biases, we propose the Rich and Fair semantic extraction network (RiFa), a novel architecture designed to not only reveal the rich semantic properties of relations, but also fairly extract relations with different annotation intensity. 
Specifically, we devise a pseudo-siamese network for independently embedding entities into separate subject and object feature vectors. The feature representation can distinguish the semantic difference between subject form and object form of the same entity, and thus avoids implicitly imposing constraints or biases on relations, which usually happens in traditional methods. Moreover, to fairly handle relations with different annotation intensities, the relations of a subject-object pair are predicted based on both the embeddings of subject-object pair and the contextual visual features extracted from the smallest bounding box covering both entities in the pair. Then, a relation score that does not depend on the statistical distribution of annotations is devised to rank all possible predictions and promote the true relations, even with a few annotations.

Extensive experiments on the Visual Genome \cite{Krishna2017Visual} dataset show that RiFa not only achieves the state-of-the-art standard recall performance on several task settings comparing to previous approaches, but also has the best overall per relation performance. The RiFa model also demonstrates significantly better performance on capturing different semantic properties. Note that our method only relies on the simple convolutional neural network, but can obtain the state-of-the-art performance, compared to those sophisticated methods such as graph neural network based ones. This means that the principle behind our design that eliminates the biases provides a new sight to the study of scene graph generation.

\section{Related Work}

There are many ways to formulate the task of extracting relations between entities in images. Visual relation detection (VRD) \cite{Lu2016} and scene graph generation \cite{Johnson2015} are two common tasks that attract the attention of many researchers recently for the challenging nature of the tasks and the complexity of the corresponding datasets used for training.

For visual relation detection, the current works are mainly focus on capturing interactions between object pairs \cite{Lu2016-2,Plummer2016,Peyre2017}.
For scene graph generation, the approaches can be divided into two types based on whether they utilize external knowledge \cite{Liao2016,Zhang2017ICCV,Wang2019}.

\textbf{Scene Graph Generation Based on Visual Features.}
The DR-Net models the scene graph generation as inference on a conditional random field (CRF) \cite{dai2017detecting}.
\cite{xu2017scenegraph} uses Recurrent Neural Network (RNN) for relation extraction and message passing for improving predictions iteratively.
\cite{Newell2017Pixels} propose a end-to-end scene graph extraction approach based on associative embeddings to predict relations using heatmap.
Qi et al propose an attentive relational network for mapping images to scene graphs \cite{Qi2019}.

To improve the scalability of models,
\cite{Li2018} tries to tackle the problem of the quadratic combinations of possible relationships by introducing a concise subgraph-based representation of the scene graph.
\cite{Yang2018} proposes a scene graph generation model based on Graph R-CNN, which uses a relation proposal network to deal with the quadratic number of potential relations between objects. Our model also use an approach that is similar in spirit to the above two for filtering subject-object pairs. 
However, our model uses a measurement learned through data to predict the likelihood of a subject-object pair having relations.

To handle the imbalanced annotations of relations, \cite{Chen2019} builds a graph representation of statistical correlations between object pairs and their relationships, and uses a graph neural network to learn the interplay between relationships and objects to generate scene graph. 

\begin{figure*}[tp!]
  \centering
  \includegraphics[width=0.8\textwidth]{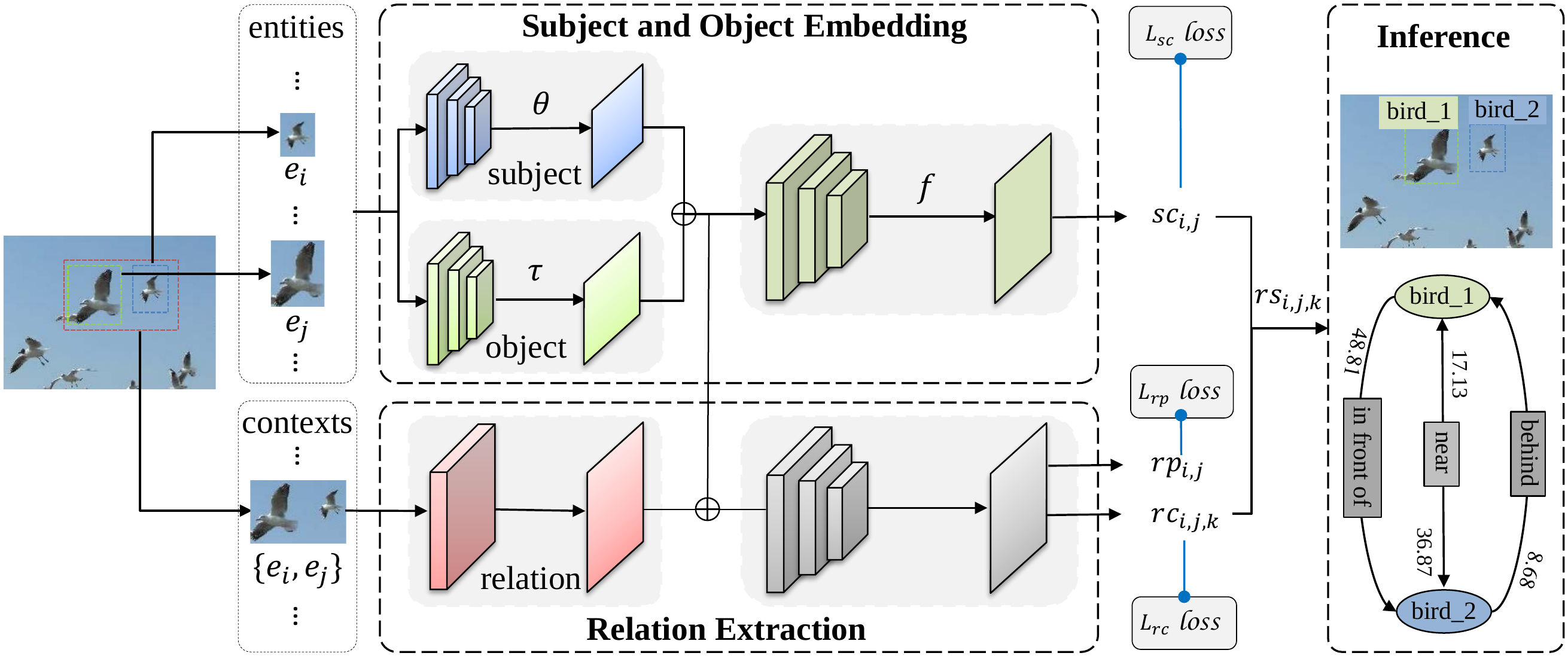}
  \caption{The overall pipeline of the RiFa model.}\label{fig:RiFa}
\end{figure*}

\textbf{Scene Graph Generation Based on External Knowledge.} \cite{Lu2016} combines visual and language features for relation detection of object pairs. \cite{Zellers2017Neural} utilizes the statistical information about repeated motifs in the datasets and global context to predict all possible relations between pairs of objects. \cite{Li2017ICCV} uses region captions to provide context for scene graph generation. \cite{Plesse2018Visual} incorporates knowledge distillation into neural network to improve the prediction. \cite{Zhang2019} joints visual features from images and semantic features from external knowledge bases to handle imbalanced distribution of triples. 

Unlike many of the above approaches that only treat the scene graph task as labeling entities and relations between entities, our model tries to extract richer semantics including relations between entities with specific semantic properties and alleviate the imbalance between different relations by eliminating biases.

\section{The RiFa Model}
In this paper, we argue that the performance bottleneck of most existing scene graph generation methods largely stems from the unexpected biases in both the unreasonable assumption of relation properties and the imbalanced relation annotations. Therefore, our goal is to generate more accurate scene graph by utilizing this clue and a simple conventional convolutional network architecture. We propose the Rich and Fair semantic extraction network (RiFa), which treats the scene graph generation as a process of extracting knowledge graph, and is able to extract richer semantics and preserve the fairness for relations with imbalanced distributions.


The overall pipeline of our RiFa model is shown in Figure \ref{fig:RiFa}. As the figure shows, given an image, the RiFa model first extracts the visual feature maps with a backbone. Then, each entity $e_i$ in the image is embedded into subject and object feature vectors $\theta(e_i)$ and $\tau(e_i)$ separately from its visual feature obtained by ROI align on feature maps, entity class, and bounding box through a pseudo-siamese network. Semantic connection strength $\mathrm{sc}_{i,j}$, the likelihood of whether a subject-object pair $(e_i, e_j)$ has relations, is predicted by passing subject and object embeddings through a three layers fully-connected neural network (FCN). The top-$N$ pairs that are most likely to have relation are passed for further relations extraction.
The relations $\mathrm{rc}_{i,j,1}, \ldots, \mathrm{rc}_{i,j,n}$ between subject-object pair $(e_i, e_j)$ as well as the possibilities $\mathrm{rp}_{i,j}$ that the pair has relations are predicted based on relation (context), subject, and object features, where $n$ is the number of relation classes. Finally, the resulting triples are sorted based on relation score $\mathrm{rs}_{i,j,k}$ computed from $\mathrm{sc}_{i,j}$, $\mathrm{rp}_{i,j}$, and $\mathrm{rc}_{i,j,k}$, and the top-$k$ triples are selected as the final result of relation extraction.

\subsection{Rich Subject and Object Semantic Embedding}
In existing scene graph generation models, they usually embed every entity $e$ into a vector through the same projection function $\sigma$. 
However, a entity has notable differences visually and semantically when it is used as subject and object in a relation. Ignoring such difference may have negative impact on capturing semantic properties of relations such as asymmetry.
To address the assumption bias, this paper converts the visual entities into semantically different subjects and objects features separately through two projection $\theta$ and $\tau$, which are learned through a pseudo-siamese network whose two branches are the same convolutional neural network without sharing parameters. In such a way, every visual entity $e$ is represented by independent subject embedding $\theta(e)$ and object embedding $\tau(e)$ simultaneously. 

\begin{figure}[tp!]
  \centering
  \includegraphics[width=0.9\columnwidth]{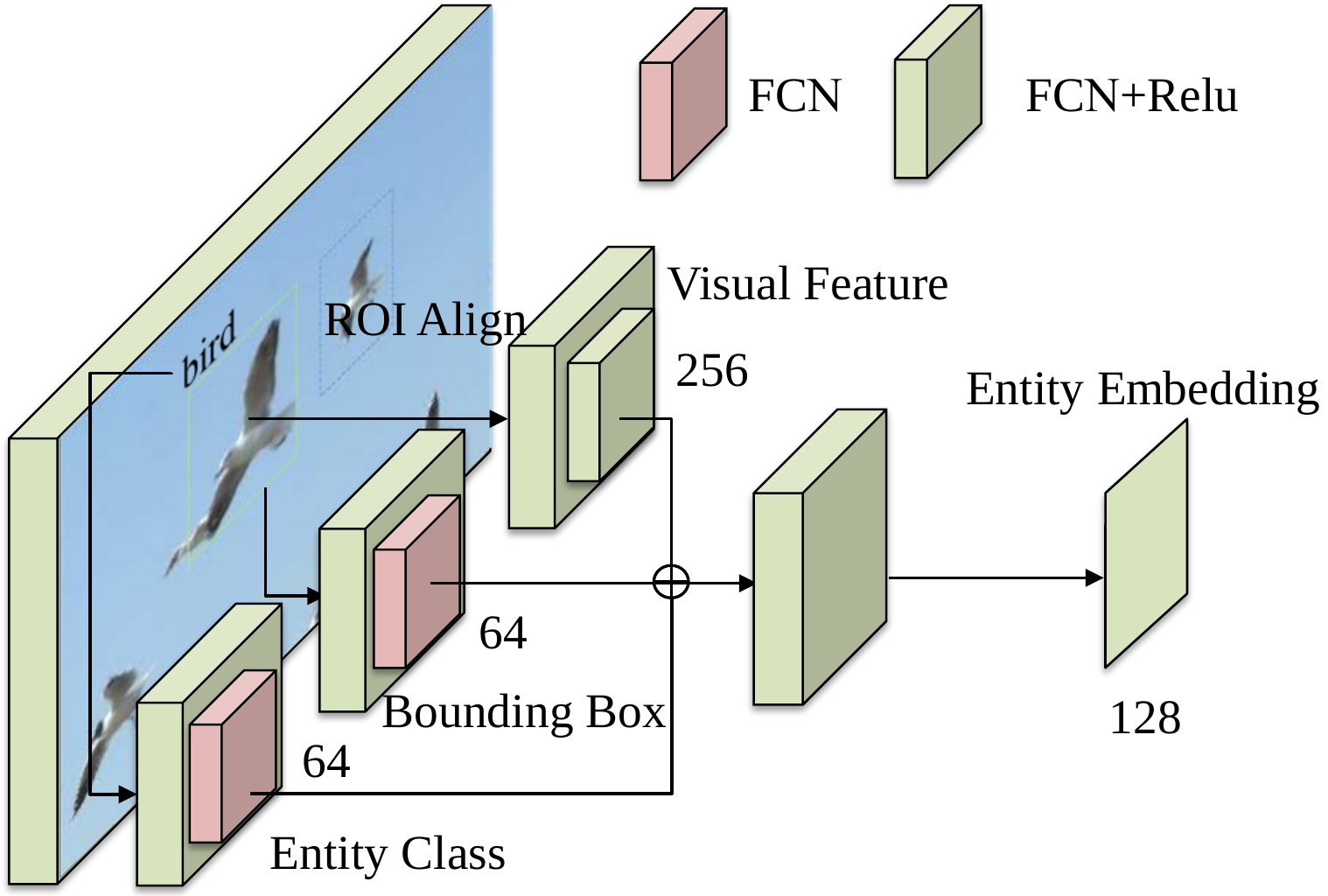}
  \caption{The network structure for entity embedding.}\label{fig:s-o-embedding}
\end{figure}

The two branches of the pseudo-siamese network for subject and object entity embedding have the same structure as the one described in Figure \ref{fig:s-o-embedding}.
The inputs of the network are one hot embedding of entity class (or the predicted class vector in case of when the entity class is not provided), entity bounding box, and feature vector obtained by ROI align \cite{ROIAlign} on visual feature maps.

In order to produce the correct subject-object pairs passed to relation prediction, we introduce a binary classifier for filtering out pairs that are unlikely to have relations, which require a metric for measuring likelihood that a subject-object pair has reasonable relations. However, common measurements such as cosine (angular) distance or Gaussian distance have both symmetry and triangle inequality properties, which may introduce bias that forces relations to become symmetric and/or transitive. For a general relation, it is not necessarily to have properties such as symmetry or transitivity. Hence, we define a new measurement named \emph{semantic connection strength}.

The semantic connection strength of a subject-object pair $(e_i,e_j)$ is defined as
\begin{equation}
\mathrm{sc}_{i,j} = f(\theta(e_i), \tau(e_j)),
\end{equation}
where $\theta(e_i)$ and $\tau(e_j)$ are the subject embedding of $e_i$ and object embedding of $e_j$, and $f$ is a function learned by passing the concatenation of subject and object embeddings through a three layers FCN with activation function $\tanh$ to map the result to $[-1, 1]$. When $\mathrm{sc}_{i,j}$ is approaching $1$, it means that the $i$th entity $e_i$ and the $j$th entity $e_j$ are more likely to have relations; otherwise, it means they are less likely to have relations. 

Because the subject-object pairs which have relations are usually only $10\%$ of all pairs, so when defining the semantic connection loss based on the classic cross-entropy, the model cannot achieve expected performance. Thus, we further introduce a \emph{semantic connection loss} $L_{sc}$ for training the pseudo-siamese network. It uses an addition penalty term to compensate the bias caused by the massive pairs that have no relations, which is defined as the average cross-entropy of all subject-object pairs that have relations. For an image containing $m$ entities, let us denote the set of indexes of subject-object pairs which have relations as $I_t$ and the remaining set as $I_f$ according to the ground truth. It is obvious that $|I_t| + |I_f| = m^2$.
Let
\begin{equation}
\delta^{C}_{i,j} = \begin{cases}
1, & (i,j) \in C\\
0, & \mbox{otherwise}\\
\end{cases}.
\end{equation}
The semantic connection loss is defined as follows.%
\begin{equation}
L_{sc} = \sum_{i,j \in [1, m]} - \lambda_{i,j} \cdot ln\left(\frac{1 - (-1)^{\delta^{I_t}_{i,j}} \mathrm{sc}_{i,j}}{2}\right),
\end{equation}
where
\begin{equation}
\lambda_{i,j} = \frac{1}{m^2} + \frac{\delta^{I_t}_{i,j}}{|I_t|}.
\end{equation}
Here $\frac{1}{m^2}$ is the weight for obtaining the classic cross-entropy, while $\frac{\delta^{I_t}_{i,j}}{|I_t|}$ is the weight for addition reward to compensate the imbalance between pairs that have relation and those do not.

With the pseudo-siamese network, for a given image with $m$ entities, we can select the subject-object pairs with top-$N$ highest semantics connection strength for further relation prediction from the $m^2$ possible combinations.

\subsection{Fair Extraction of Imbalanced Relations}
As aforementioned, in practice the annotations for different relations are quite imbalanced. For example, in the Visual Genome dataset about $76\%$ of the ground truth annotations are triples about only $5$ relations: ``wearing, has, on, of, in''. As a result, the relations that frequently occur in images are easy to extract and dominates the model predictions with a strong bias, while the relations that only occur in a small number of images are hard to extract, especially for frequency based approaches. In order to make the relation extraction more fair for different types of relations, we predict relations between two entities by combining their subject and object entity embedding with relation embedding, and the possibility that the subject and object entities have relations simultaneously.

In order to define the relation embedding, we first define the relation bounding box for a relation instance as the smallest rectangle which covers the bounding boxes of both its subject and object. The relation embedding is the visual feature within the relation bounding box in the image, which describes the context of relations between the two entities.
Besides, the combination of relation embedding with the subject and object entity embeddings can also help to solve the problem that multiple subject-object pairs have overlapping relation bounding boxes. It also support the case that a single subject-object pair has multiple relations.

To accurately filter out subject-object pairs which do not have relations, the \emph{relation possibility} $\mathrm{rp}_{i,j}$ is predicted simultaneously with the relation classes to measure the possibility that the $i$th subject and the $j$th object have relations. We introduce \emph{relation possibility loss} $L_{rp}$ to guide the learning of the relation extraction network, in which a penalty term similar to the one defined in $L_{sc}$ is used.

Formally, for the $N$ subject-object pairs proposed for further prediction, let us denote the index set of subject-object pairs as $E$, those have relations as $E_t$ and the remaining set as $E_f$ according to the ground truth. We have $|E_t| + |E_f| = N$ and $|E_t| \leq |I_t|$. The relation possibility loss $L_{rp}$ is defined as follows.%
\begin{equation}
L_{rp} = \sum_{(i,j) \in E} -\gamma_{i,j} \cdot ln\left(\mathrm{rp}_{i,j}^{\delta^{E_t}_{i,j}} (1-\mathrm{rp}_{i,j})^{(1-\delta^{E_t}_{i,j})} \right),
\end{equation}
where
\begin{equation}
\gamma_{i,j} = \dfrac{1}{N} + \dfrac{\delta^{E_t}_{i,j}}{|E_t|}.
\end{equation}
Similar to the case of $\lambda_{i,j}$, $\frac{1}{N}$ is the weight for obtaining the classic cross-entropy, while $\frac{\delta^{E_t}_{i,j}}{|E_t|}$ is the penalty term.



\subsection{Training and Inference}

\noindent\textbf{Training}. The final module of the RiFa model is to predict relations between subject-object pairs. For the training of relation prediction, we define the \emph{relation class loss} $L_{rc}$, in which we only need to consider the subject-object pairs that have relations. So the relation class loss can be defined as the cross-entropy of all relations in the subject-object pairs which have relations.
Let us denote the index set of relations between the $i$th subject and the $j$th object as $R_{i,j}$ and $r = \sum_{(i,j) \in I_t} |R_{i,j}|$. Then, since there can be more than one relation between the same subject-object pair, $r \geq |I_t|$. The relation class loss $L_{rc}$ is defined as follows.%
\begin{equation}
L_{rc} = \frac{1}{r} \sum_{(i,j) \in I_t} \sum_{k \in R_{i,j}} -ln(\mathrm{rc}_{i,j,k}),
\end{equation}
where $\mathrm{rc}_{i,j,k}$ is the possibility that the relation between the $i$th subject and the $j$th object is the $k$th relation.

The loss function for RiFa is defined as
\begin{equation}
L = L_{sc} + L_{rp} + L_{rc},
\end{equation}
where the above three losses are weighted equally.

\noindent\textbf{Inference}. Based on the above description of the RiFa model, three kind of values can be produced in the inference: semantic connection strength $\mathrm{sc}_{i,j}$, relations possibility $\mathrm{rp}_{i,j}$, and relation classes map $\mathrm{rc}_{i,j}$ which constitutes by the possibility of the relation between $i$th subject and $j$th object belonging to each relation classes. Based on these values, we propose the following \emph{relation score} $\mathrm{rs}_{i,j,k}$ to systematically represent the likelihood that the $i$th subject and the $j$th object have the $k$th relation.
\begin{equation}
\mathrm{rs}_{i,j,k} = \mathrm{sc}_{i,j} + \beta \cdot \mathrm{rp}_{i,j}^2 \cdot \mathrm{rc}_{i,j,k},
\end{equation}
where $\beta$ is a parameter for tuning the weight of different scores. The semantic connection strength is the core part of the relation score, which distinguishes subject-object pairs that do have relations with those do not. The relation possibility can be viewed as a refinement of the semantic connection strength, which more accurately reflects the possibility that a subject-object pair have relations and can alleviate the negative impact of the false-positives in the top-$N$ subject-object pairs. It can also benefit subject-object pairs that have multiple relations by lowing the ranks of triples that are not instances of relations and increasing the ranks of triples that reflect different relations between the same subject and object. Because it is very common in datasets such Visual Genome that there are multiple relations between the same subject-object pair, it can increase the chance that relations with only a few annotations to be extracted with a higher relation scores.

The advantage of this definition is that it does not depend on statistical distribution of relation instances. So both relations with a large number of annotations and those with only a few have relatively equal chance to be ranked in the top-$k$ predictions. 

\noindent\textbf{Implementation}. We implement and train the RiFa in TensorFlow \cite{Abadi2016TensorFlow}. RiFa takes a 512x512 image for input and outputs a vector to indicate the predicted relation. The visual feature maps of images are extracted first, which contain visual features of all entities and relations between entities. Because the size of entities in images is variant, it is difficult to capture the visual and semantic features of entities with different size. Thus, to archive precise object classification and relation extraction, we draw inspiration from Feature Pyramid Network (FPN) \cite{FPN} to generate both $32 \times 32$ and $16 \times 16$ feature maps to obtain visual information of different granularity after extracting visual features through a VGG19 network. Unlike the FPN model which utilize different feature map separately, we combine them together for future usage.


\section{Experiments and Evaluations}

\subsection{Experiment Settings}
All the experiments are performed on a workstation with two GeForce GTX 1080Ti GPUs, which has Ubuntu 18.04 operation system, CUDA 9.0, CUDNN 7.0, and TensorFlow 1.8. For the hyper-parameters, we set 
$\beta = 120$ and $N = 100$ across all experiments.

\textbf{Dataset.} The performance of RiFa is evaluated on Visual Genome, using the publicly available preprocessed version with $150$ classes and $50$ relations (VG150) and the same split by \cite{xu2017scenegraph}.

\textbf{Task Setting.} The task of scene graph generation is to produce a set of triples of form $(s, p, o)$, where the subject $s$ is an entity in the image defined by its bounding box, the object $o$ can be an entity or a class. In case that $o$ is a class, $(s, p, o)$ represents that the entity $s$ is an instance of class $o$; otherwise, $p$ is a relation between two entities in the image. A triple is correct according to the ground truth, if
the classes of entities and relations between entities match those of ground truth.
Following \cite{xu2017scenegraph} and \cite{Newell2017Pixels}, this paper uses the standard evaluation metric R@$k$ for scene graphs, which measures the percentage of ground truth triples in the set of top-$k$ proposals. The performance of RiFa on the following three tasks are evaluated: 1) \textbf{PredCls:} given an image, bounding boxes, and classes of entities defined by the bounding boxes, predicting relations between entities; 2) \textbf{SGCls:} given an image and bounding boxes for entities in the image, predicting the classes of entities and relations between entities; 3) \textbf{SGGen:} given an image only, predicting the classes of entities and relations between entities. 

\subsection{Comparison to State of the Art}
We first evaluate our RiFa on the Visual Genome dataset with comparison to state-of-the-art methods, including VRD \cite{Lu2016}, IMP \cite{xu2017scenegraph}, Pixel2Graph \cite{Newell2017Pixels}, GPSKD \cite{Plesse2018Visual}, MotifNet \cite{Zellers2017Neural}, Graph R-CNN \cite{Yang2018}, LSVRU \cite{Zhang2019}, and CISC \cite{Wang2019}. In order to perform the SGCls task with RiFa, we extend the model of RiFa with an auxiliary network for entity classification after predicting the relations between a subject-object pair. The entities are classified not only based on the their own embeddings but also relations between entities. During the evaluation, both SGCls and PredCls tasks are conducted with RiFa alone. 
The comparison results on the above three tasks are listed in Table \ref{tab:results}.

\begin{table}[tb!]
\centering
\begin{tabular}{lrrrr}
  \toprule
    & \multicolumn{2}{c}{SGCls} & \multicolumn{2}{c}{PredCls}\\
    & R@50 & R@100 & R@50 & R@100 \\
  \midrule
  VRD & 11.80 & 14.10 & 27.90 & 35.00 \\
  IMP & 21.72 & 24.38 & 44.80 & 53.00 \\
  Pixel2Graph & 26.5 & 30.0 & 68.0 & 75.2  \\
  GPSKD & 35.55 & 42.74 & 67.71 & 77.60 \\
  MotifNet & 35.8 & 36.5 & 65.2 & 67.1 \\
  Graph R-CNN & 29.6 & 31.6 & 54.2 & 59.1 \\
  LSVRU & 36.7 & 36.7 & 68.4 & 68.4\\
  CISC & 27.8 & 29.5 & 53.2 & 57.9\\
  \midrule
  RiFa & \textbf{37.62} & \textbf{44.38} & \textbf{80.64} & \textbf{88.35}\\
  \bottomrule
\end{tabular}
\caption{Comparison results for SGCls and PredCls tasks.}
\label{tab:results}
\end{table}

For the R@100 metric, RiFa archives the state-of-the-art performance on the PredCls and SGCls tasks. For the R@50 metric, RiFa archives the best performance on the PredCls task and has a performance comparable to that of state-of-the-art approaches on the SGCls task.

\begin{table}[b!]
\centering
\begin{tabular}{lrr}
  \toprule
    & R@50 & R@100 \\
  \midrule
  VRD & 0.30 & 0.50 \\
  IMP & 3.40 & 4.20 \\
  Pixel2Graph & 6.70 & 7.80 \\
  MotifNet & 27.2 & 30.3 \\
  Graph R-CNN & 11.4 & 13.7 \\
  LSVRU & \textbf{27.9} & \textbf{32.5}\\
  CISC & 11.4 & 13.9\\
  \midrule
  RiFa & 20.86 & 26.68 \\
  \bottomrule
\end{tabular}
\caption{Comparison results for the SGGen task.}
\label{tab:sggen}
\end{table}

For the SGGen task, we report the results in Table \ref{tab:sggen} for the sake of completeness by combining RiFa with bounding boxes and entity classes detected by a Faster R-CNN \cite{FRCNN} with ResNet-101 \cite{ResNet}, which is pre-trained with the COCO dataset \cite{COCO} and fine-tuned on the Visual Genome dataset. RiFa still has a reasonably good performance on it comparing to other approaches. We believe RiFa can perform better with a better object detector and more fine-tuning. The models in \cite{Zellers2017Neural} and \cite{Zhang2019} achieve a better performance mainly because they either utilize the statistical clue from the dataset or introduce external knowledge base. In models that generate scene graphs without such prior information, e.g. Pixel2Grpah, Graph R-CNN, CISC, our RiFa model achieves better performance for both $R@50$ and $R@100$.

\subsection{Ablation Study}
We explore $3$ settings of the RiFa models to verify the effectiveness of the network for relation prediction: 1) the one without relation embedding (RE), 2) the one without subject and object embeddings (SOE), 3) the one without relation possibility (RP). The intuition of concatenating relation embedding with subject and object embeddings is that the latter contains both visual and semantic features of the entities in the subject and object, while the former provides the context information about the relation between these two entities, and these two features are complementary.

As shown in Table \ref{tab:ablation},
\begin{table}[tb!]
\centering
\begin{tabular}{lrrrr}
  \toprule
    & \multicolumn{2}{c}{PredCls}\\
    & R@50 & R@100 \\
  \midrule
  RiFa & \textbf{80.64} & \textbf{88.35}\\
  RiFa w/o RE & 60.43 & 71.33 \\
  RiFa w/o SOE & 45.12 & 61.84 \\
  RiFa w/o RP & 77.32 & 85.88 \\
  \bottomrule
\end{tabular}
\caption{Ablation study of RiFa.}
\label{tab:ablation}
\end{table}
the combination of relation embedding with subject and object embeddings outperforms models that with only one feature by a large margin, which demonstrates the efficacy of the concatenation. The relation possibility is designed to reduce false-positives by lowering the relation score of entities that actually have no relations. The experiment result indicates that by incorporating relation possibility in to the relation extraction network, the performance is clearly improved.


\subsection{Unbiased Scene Graph Generation}

\subsubsection{Performance on Capturing Semantic Properties}

We conduct experiments on comparing the ability of different neural networks on capturing asymmetry, symmetry, and inverse properties of relation, because they are the most common properties for relations in Visual Genome. The asymmetric relations ``wearing, has, riding, on, holding, parked on, walking on'', symmetric relations ``and, near'', and mutually inverse relation pairs ``(in front of, behind), (above, under), (has, part of)'' are chosen for the evaluation. Let us denote the set of top-$k$ predictions for the PredCls task as $P_k$. We modify the definition of whether a triple in $P_k$ matches the ground truth for the three groups of relations as follows.

\textbf{Asymmetry.} Let $p$ be an asymmetric relation. If $(s,p,o) \in P_k$ matches the ground truth and $(o,p,s) \not\in P_k$, then $(s,p,o)$ matches the ground truth w.r.t. asymmetry property.

\textbf{Symmetry.} Let $p$ be a symmetric relation. If $(s,p,o) \in P_k$ matches the ground truth and $(o,p,s) \in P_k$, then $(s,p,o)$ matches the ground truth w.r.t. symmetry property. 

\textbf{Inverse.} Let $p$ and $q$ are inverse to each other. If $(s,p,o) \in P_k$ matches the ground truth and $(o,q,s) \in P_k$, then $(s,p,o)$ matches the ground truth w.r.t. inverse property. 

The recalls for the three groups of asymmetric, symmetric, and mutually inverse relations based on corresponding matches defined in above are denoted as $R_{A}@k$, $R_{S}@k$, and $R_{I}@k$ respectively.
The three recalls of Pixel2Graph and RiFa on VG150 are listed in Table \ref{tab:res-asym}.
\begin{table}[tb!]
    \centering
    \begin{tabular}{lrrrrrr}
    \toprule
    & Pixel2Graph & RiFa \\
    \midrule
    $R_{A}@50$ & 65.33 & \textbf{84.41}\\
    $R_{A}@100$ & 53.02 & \textbf{84.34}\\
    $R_{S}@50$ & \textbf{35.84} & 15.53\\
    $R_{S}@100$ & \textbf{48.92} & 29.65\\
    $R_{I}@50$ & \textbf{11.48} & 8.05\\
    $R_{I}@100$ & \textbf{29.34} & 22.16\\
    \bottomrule
    \end{tabular}
    \caption{The performance of Pixel2Graph and RiFa on capturing asymmetry, symmetry, and inverse properties.}
    \label{tab:res-asym}
\end{table}

We can see that the Pixel2Graph model has low $R_{A}@k$ values and unusually high $R_{S}@k$ values, which indicates that it may contain bias that prefers to treat relations as symmetric. As a result, the Pixel2Graph model does not perform well on capturing asymmetry property of relations.
The RiFa model has low $R_{S}@k$ and $R_{I}@k$ values. By analyzing the annotations of VG150, we find that there are only a few annotations that reflect the symmetry and inverse properties, which may be the cause of the low performance of RiFa on symmetric and mutually inverse relations.

\begin{table}[b!]
    \centering
    \begin{tabular}{clrrrrrr}
    \toprule
    & & Pixel2Graph & RiFa \\
    \midrule
    \multirow{4}{*}{\rotatebox{90}{VG+50\%}}
    & $R_{S}@50$ & \textbf{43.72} & 33.47\\
    & $R_{S}@100$ & \textbf{58.11} & 52.10\\
    & $R_{I}@50$ & 58.59 & \textbf{61.31}\\
    & $R_{I}@100$ & 72.96 & \textbf{77.62}\\ \midrule
    \multirow{4}{*}{\rotatebox{90}{VG+100\%}}
    & $R_{S}@50$ & \textbf{44.86} & 36.44\\
    & $R_{S}@100$ & \textbf{59.39} & 56.04\\
    & $R_{I}@50$ & 62.56 & \textbf{69.48}\\
    & $R_{I}@100$ & 73.47 & \textbf{81.34}\\
    \bottomrule
    \end{tabular}
    \caption{The performance after extending the dataset.}
    \label{tab:res-ext}
\end{table}

To verify this, we construct the VG+50\%/VG+100\% dataset by automatically extending 50\%/100\% annotations about the above chosen symmetric and mutually inverse relations with their symmetric or inverse forms. For instance, $(o,p,s)$ is added as a new annotation if $(s,p,o)$ is a annotation and $p$ is symmetric, and $(o,q,s)$ is added as a new annotation if $(s,p,o)$ is a annotation and $q$ is inverse to $p$.

As shown in Table \ref{tab:res-ext}, on these two extended datasets, the $R_{S}@k$ and $R_{I}@k$ values of RiFa model become comparable to or even significantly higher than those of Pixel2Graph model, which provides support to our conjecture.

With the addition annotations, RiFa performs significantly better than Pixel2Graph on capturing asymmetry and inverse properties and has comparable performance on capturing symmetry property despite the advantage given by the bias of Pixel2Graph model.

\subsubsection{Performance on Handling Imbalanced Relations}

There are 34 relations in VG150 that have a very small number of annotations (less than $0.5\%$ of all annotations).
\begin{table}[tb!]
    \centering
    \scalebox{0.8}{
    \begin{tabular}{lr|lr|lr}
    \toprule
    \multicolumn{2}{c|}{RiFa} & \multicolumn{2}{c|}{Pixel2Graph} & \multicolumn{2}{c}{MotifNet}\\
    Relation & R@100 & Relation & R@100 & Relation & R@100\\
    \midrule
    wearing & 96.78 & on & 86.15 & wearing & 97.55\\
    has & 93.05 & wearing & 86.14 & on & 95.18 \\
    riding & 92.75 & has & 85.10 & has & 93.57 \\
    on & 92.19 & of & 81.03 & of & 90.36 \\
    of & 89.45 & riding & 78.26 & riding & 89.70 \\
    holding & 87.81 & holding & 77.73 & holding & 83.96 \\
    \underline{parked on} & 86.64 & wears & 77.62 & wears & 80.72 \\
    \underline{walking on} & 83.46 & in & 73.13 & in & 79.70 \\
    in & 80.92 & near & 69.51 & \underline{walking on} & 75.89 \\
    sitting on & 76.33 & with & 68.86 & sitting on & 70.42 \\
    wears & 74.41 & above & 68.76 & behind & 68.89 \\
    \underline{eating} & 71.89 & sitting on & 68.15 & near & 67.94 \\
    \underline{carrying} & 70.51 & carrying & 65.91 & \underline{carrying} & 67.91 \\
    \underline{laying on} & 68.56 & behind & 65.39 & with & 66.38 \\
    \underline{at} & 66.22 & \underline{under} & 64.52 & \underline{parked on} & 64.30 \\
    \bottomrule
    \end{tabular}}
    \caption{Top-$15$ per relation performance. Relations with a very small number of annotations are labeled with underline.}
    \label{tab:ppr-top}
\end{table}

\begin{figure*}[tb]
\centering
\includegraphics[width=\textwidth]{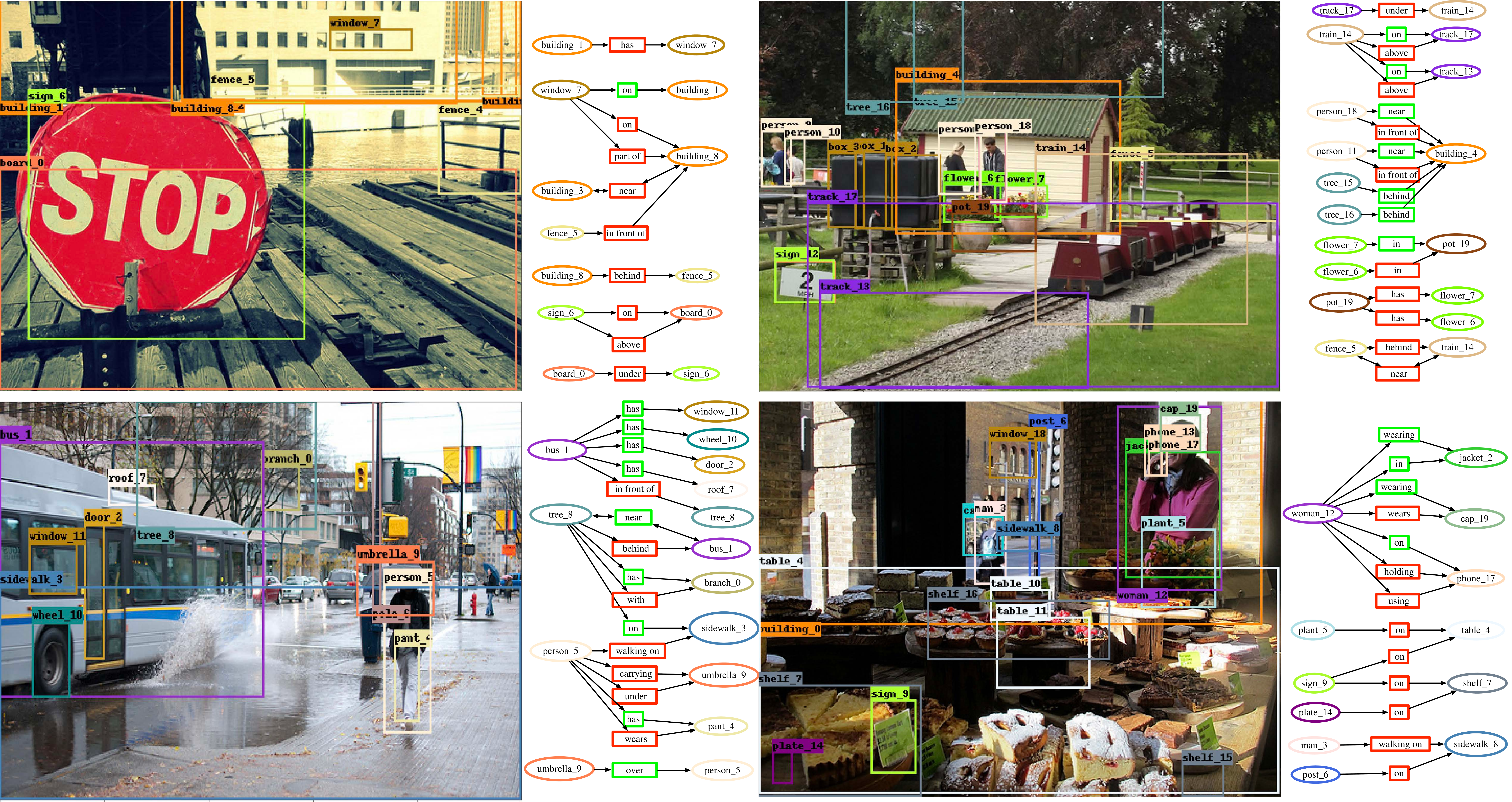}
\caption{Qualitative examples of relation prediction. Relations outlined in green correspond to triples that match ground truth. Relations in red are correct instances that are not been annotated in the VG150 dataset.}
\label{fig:example}
\end{figure*}

\begin{figure}[b!]
    \centering
    \includegraphics[width=0.8\columnwidth]{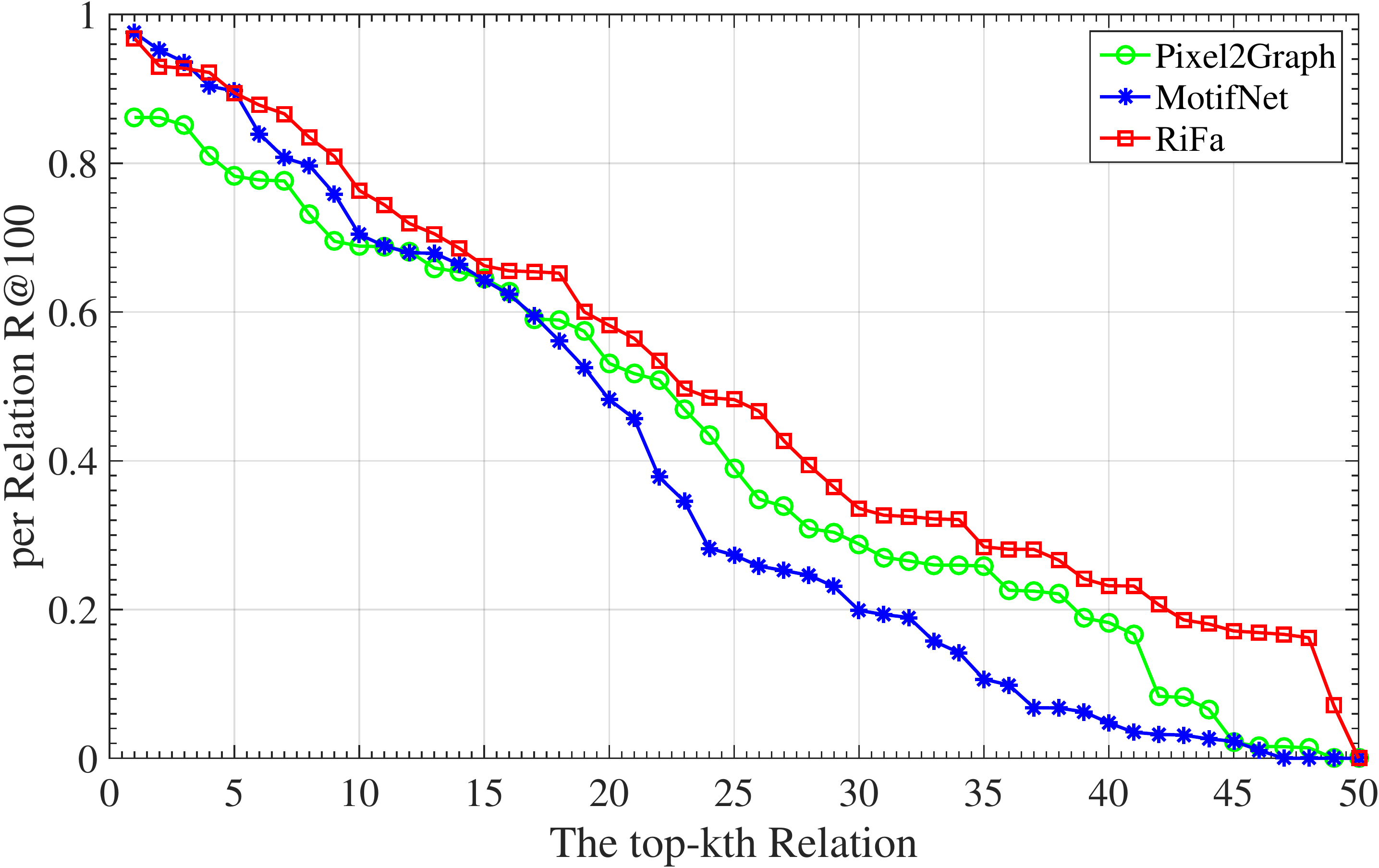}
    \caption{The per relation $R@100$ values of the $50$ relations sorted in descendent order.}
    \label{fig:perrel}
\end{figure}

From Table \ref{tab:ppr-top}, we can see that the top-$15$ list of RiFa contains more relations ($40\%$) with a small number of annotations, which indicates that the RiFa performs better on learning to predict not only relations with a large number of annotations but also relations with only a few annotations.

In Figure \ref{fig:perrel}, we can see that the Pixel2Graph model has the smallest number of relations whose per relation $R@100$ values are higher than $80\%$. While the MotifNet model leads to lower per relation recall for a large number of relations which are mostly relations with relatively small number of annotations. The overall per relation recalls of our RiFa model are consistently better than the other two models. In addition, the mean per relation recall $R@100$ of all $50$ relations is $41.61\%$ for the Pixel2Graph model, $37.79\%$ for the MotifNet model, and $48.86\%$ for the RiFa model. Hence, both evidences indicate that our RiFa model has better per relation performance and is more fair for relations with different scale of annotations.

\subsection{Qualitative Results}

As shown in Figure \ref{fig:example}, 
our RiFa model is able to extract relations with only a few annotations, such as ``walking on'', ``carrying'', ``over'', ``using''. It also demonstrates the ability to capture semantic properties of relations such as asymmetry, symmetry and invert. For example, ``walking on'' is always predicted to be asymmetric. The two triples (tree\_8, near, bus\_1) and (bus\_1, near, tree\_8) are both predicted for the bottom left image, which is a reflection that ``near'' relation is symmetric. Mutually inverse relation pairs, such as (has, part of), (in front of, behind), (above, under), are also been correctly recognized by RiFa.

\section{Conclusion}

This paper presented a new model design principle for improving the performance of scene graph generation, i.e. avoiding two types of biases in models: assumptions about semantic properties of relations such as symmetry and imbalanced annotations for different relations. Based on this principle, the RiFa model was proposed to not only capture rich semantic properties of relations, but also fairly extract relations with different annotation intensity. 
Experimental results shown that the RiFa model achieved the state-of-the-art performance on several tasks comparing to other sophisticated methods, even though it only relies on simple convolutional neural network, which demonstrated the effectiveness of the proposed design principle.

For the future work, we shall further verify the generality of the design principle on other relation extraction models. The preliminary experiments in this paper already shown that many trained models did not actually capture the rich semantics of images. It is also very interesting to design metrics to evaluate the capability of each model on capturing different semantics and help explain which kind of semantics is captured in the generated scene graphs.

{\small
\bibliographystyle{ieee_fullname}
\bibliography{egbib}
}

\end{document}